\documentclass[x11names]{article}

\usepackage{microtype}
\usepackage{graphicx}
\usepackage{booktabs} 

\usepackage{hyperref}

\usepackage[accepted]{icml2026}

\usepackage{amsmath}
\usepackage{amssymb}
\usepackage{mathtools}
\usepackage{amsthm}

\usepackage{graphicx} 
\usepackage{booktabs}       
\usepackage{nicefrac}       
\usepackage{microtype}      
\usepackage{xcolor}         
\usepackage{pifont}
\usepackage{algorithm}
\usepackage{algorithmic}
\usepackage{colortbl}
\usepackage{multirow}
\usepackage{rotating}
\usepackage{ulem}
\usepackage{subcaption}

\newcommand{\retop}[3]{\underset{#1 \in #2}{\operatorname{Ret}_{#3}}}

\DeclareMathOperator*{\argmax}{arg\,max}
\DeclareMathOperator*{\argmin}{arg\,min}

\newif\ifproofread

\usepackage[capitalize,noabbrev]{cleveref}

\theoremstyle{plain}

\theoremstyle{definition}

\theoremstyle{remark}

\usepackage[textsize=tiny]{todonotes}

\icmltitlerunning{Zero-shot Concept Bottleneck Models}

\begin{document}

\twocolumn[
\icmltitle{Zero-shot Concept Bottleneck Models}




\begin{icmlauthorlist}
\icmlauthor{Shin'ya Yamaguchi}{ntt,kyoto}
\icmlauthor{Kosuke Nishida}{ntt}
\icmlauthor{Daiki Chijiwa}{ntt}
\icmlauthor{Yasutoshi Ida}{ntt}
\end{icmlauthorlist}

\icmlaffiliation{ntt}{NTT}
\icmlaffiliation{kyoto}{Kyoto University}

\icmlcorrespondingauthor{Shin'ya Yamaguchi}{shinya.yamaguchi@ntt.com}


\vskip 0.3in
]



\printAffiliationsAndNotice{}  

\begin{abstract}
    Concept bottleneck models (CBMs) are inherently interpretable and intervenable neural network models, which explain their final label prediction by the intermediate prediction of high-level semantic \textit{concepts}.
    However, they require target task training to learn input-to-concept and concept-to-label mappings, incurring target dataset collections and training resources.
    In this paper, we present \textit{zero-shot concept bottleneck models} (Z-CBMs), which predict concepts and labels in a fully zero-shot manner without training neural networks.
    Z-CBMs utilize a large-scale concept bank, which is composed of millions of vocabulary extracted from the web, to describe arbitrary input in various domains.
    For the input-to-concept mapping, we introduce \textit{concept retrieval}, which dynamically finds input-related concepts by the cross-modal search on the concept bank.
    In the concept-to-label inference, we apply \textit{concept regression} to select essential concepts from the retrieved concepts by sparse linear regression.
    Through extensive experiments, we confirm that our Z-CBMs provide interpretable and intervenable concepts without any additional training.
    Code will be available at \url{https://github.com/yshinya6/zcbm}.
\end{abstract}

\proofreadtrue

\begin{figure*}[t]
    \vspace{-3mm}
    \centering
    \centering
    \begin{minipage}[b]{0.49\linewidth}
      \centering
      \includegraphics[width=\linewidth]{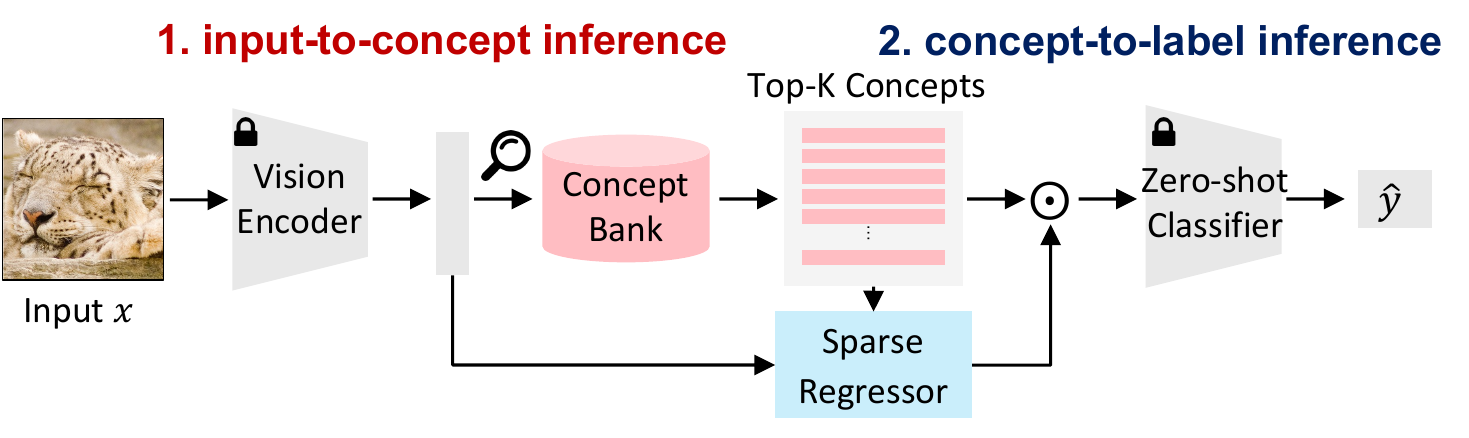}
      \subcaption{Overview of Z-CBMs}\label{fig:top_zcbm}
    \end{minipage}
    \hfill
    \begin{minipage}[b]{0.49\linewidth}
      \centering
      \includegraphics[width=\linewidth]{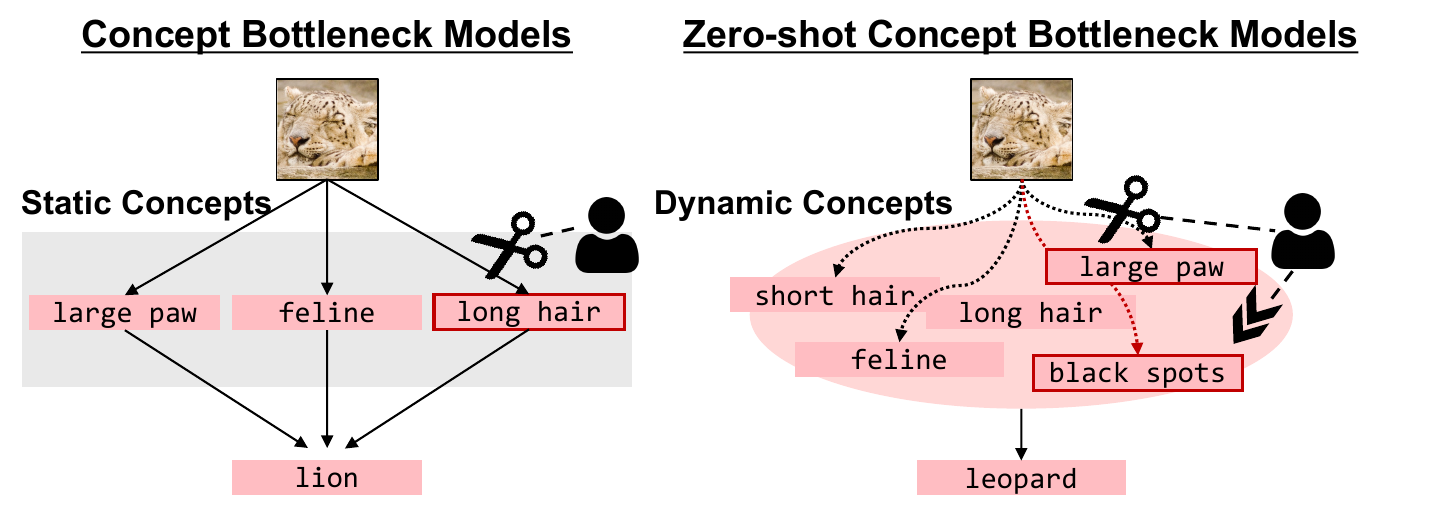}
      \subcaption{Comparison of Intervention}\label{fig:top_intervention}
    \end{minipage}
    \caption{
    Zero-shot concept bottleneck models (Z-CBMs). 
    (a) Z-CBMs predict concepts for input by retrieving them from a large-scale concept bank. Then, Z-CBMs predict labels based on the weighted sum of the retrieved concept vectors with importance weights yielded by sparse linear regression.
    (b) Z-CBMs can be intervened through modifying retrieved concepts and adding arbitrary concepts, whereas conventional CBMs allow only interventions in static concepts. The diverse and dynamic concepts produce accurate predictions and flexible collaborations with human experts.
    }
    \label{fig:top}
    \vspace{-3mm}
\end{figure*}
\section{Introduction}\label{sec:introduction}
Developing human-interpretable models remains a primary interest within the deep learning research community.
Concept bottleneck model (CBM, \cite{Koh_ICML20_concept_bottleneck}) is an inherently interpretable neural network model, which aims to explain its final prediction via the intermediate \textit{concept} predictions.
Typically, CBMs are trained end-to-end on a target task to learn the input-to-concept and concept-to-class mappings.
A concept consists of high-level semantic vocabulary for describing objects of interest in input data.
For instance, CBMs can predict the final label ``apple" from the linear combination of the concepts ``red sphere," "green leaf," and ``glossy surface."
These intermediate concept predictions not only provide interpretability but also intervenability in the final prediction by editing the predicted concepts.\looseness-1

In original CBMs~\cite{Koh_ICML20_concept_bottleneck}, a concept set for each class label is defined by manual annotations, incurring massive labeling costs greater than ones of the class labels.
To reduce these costs, Oikarinen~et~al{.}~\cite{Oikarinen_ICLR23_label-free_CBMs} and Yuksekgonoul~et~al{.}~\cite{Yuksekgonul_ICLR23_post-hoc_CBMs} automatically generate the concept sets by large language models (LLMs, e.g., GPT-3~\cite{brown_NIPS20_gpt3}) and use the multi-modal embedding space of vision-language models (VLMs, e.g., CLIP~\cite{Radford_ICML21_CLIP}) to learn the input-to-concept mapping through similarities in the multi-modal feature space.
Although modern CBMs are free from manually pre-defined concepts, their practicality is still restricted by the requirements of training input-to-concept and concept-to-class mappings on target datasets.
This means that CBMs have not been available without manually collecting target datasets and additional training of model parameters on them so far.
Furthermore, CBMs allow interventions only in static concepts that are used in training, preventing human experts from flexible interactions with arbitrary concepts.

To overcome these limitations, this paper introduces a novel problem setting of CBMs in a zero-shot manner for target tasks.
In this setting, we can access pre-trained VLMs, but we cannot know the concepts composing target data in advance.
This setting necessitates a two-stage zero-shot inference of input-to-concept and concept-to-class for unseen input samples.
The zero-shot input-to-concept inference can not be solved by a na\"ive application of VLMs as the ordinary zero-shot classification of input-to-label, because it requires identifying a subset of relevant concepts from the large set of all concepts, rather than predicting a single label.
Furthermore, the zero-shot concept-to-class inference is difficult because how to obtain the concept-to-class mapping is not obvious without target data and training, which are unavailable in this setting.
Therefore, our primary research question is: \textit{How can we achieve interpretable and intervenable concept-based prediction through zero-shot input-to-concept and concept-to-class inference without relying on target datasets and additional training?}\looseness-1

We present a novel CBM class called \textit{zero-shot concept bottleneck models} (Z-CBMs).
Z-CBMs are zero-shot interpretable models that employ off-the-shelf pre-trained VLMs with frozen weights as the backbone (Fig.~\ref{fig:top}).
Conceptually, Z-CBMs first perform \textbf{concept retrieval} to dynamically identify input-related concepts from a broad concept bank and then \textbf{concept regression} predicts the final label by simulating zero-shot classification capabilities of black-box VLMs via reconstructing the original input embedding from the retrieved concept embeddings.
Our primary contribution is to achieve zero-shot input-to-concept and concept-to-class inference with this framework without additional training.
Furthermore, our Z-CBMs allow interventions by arbitrary concepts described in natural language through the VLM feature spaces. \looseness-1

We implement the components of Z-CBMs with simple yet carefully designed and effective techniques.
For concept retrieval, Z-CBMs should cover broad domains to provide sufficient concepts for unseen inputs.
To cover broad concepts, we build a large-scale concept bank, which is composed of millions of vocabulary extracted from large-scale text caption datasets such as YFCC~\cite{Thomee_ACM16_yfcc100m}.
Given an input sample, Z-CBMs dynamically retrieve concept candidates from the concept bank with an efficient and scalable cross-modal search algorithm.
For concept regression, Z-CBMs estimate the importance of concepts for the input feature and then predict labels by the importance-weighted concept features.
However, many of the retrieved concept candidates semantically overlap with each other, and thus, the semantically duplicated concepts with high importance by a na\"ive estimation method can harm the interpretability and intervenability for humans.
To overcome this challenge, Z-CBMs find essential and mutually exclusive concepts for the final label prediction by leveraging sparse linear regression (e.g., lasso) to reconstruct the input visual feature vector by a weighted sum of the concept candidate vectors.
Combining concept retrieval and concept regression enables Z-CBMs to predict final class labels with interpretable concepts for various domain inputs without any target training.

Our extensive experiments on 12 datasets demonstrate that Z-CBMs can provide interpretable and intervenable concepts without any additional training.
Specifically, we demonstrate that the sparse concepts identified by Z-CBMs exhibit strong correlation with input images and cover the annotated concepts in existing training-based CBMs.
Furthermore, the Z-CBMs' performance can be enhanced by human intervention in the predicted concepts, emphasizing the reliability of the concept-based prediction.
We also show that Z-CBMs competitively perform with black box VLMs and existing CBMs with training.
These results suggest the practicality of Z-CBMs for various domains.\looseness-1

\section{Preliminaries}
\subsection{Concept Bottleneck Models (CBMs)}
A CBM~\cite{Koh_ICML20_concept_bottleneck} is a classifier composed of a concept predictor \(g:\mathcal{X}\to\mathcal{C}^K\) and a class label predictor \(h:\mathcal{C}^K\to\mathcal{Y}\), where \(\mathcal{X},\mathcal{C},\mathcal{Y}\) are input, concept, and class label spaces, and \(K\) is the number of concepts.
The goal is to predict the final class label \(y\in\mathcal{Y}\) of input \(x\in\mathcal{X}\) based on $K$ interpretable concepts \(C=\{c_i\in\mathcal{C}\}_{i=1}^{K}\).
To guarantee the interpretability and classification performance, \(g\) and \(h\) are jointly optimized on the following objective function~\cite{Koh_ICML20_concept_bottleneck}:
\begin{equation}\label{eq:cbm_obj}
    \min_{g, h} \mathbb{E}_{(x,C,y)\in\mathcal{D}}\left[ \mathcal{L}(h\circ g(x),y) + \alpha\sum_{i}^K\mathcal{L}(g(x)_i,c_i)\right],
\end{equation}
where \(\mathcal{D}\) is a training dataset, \(\alpha\) is a hyperparameter, and \(\mathcal{L}\) is a supervised loss function such as softmax cross-entropy loss.
That is, CBMs' interpretability is defined by their ability to detect concepts in the input accurately, which is obtained through the training of input-to-concept and concept-to-class predictions.
In this sense, CBMs have the challenge of requiring human annotations of concept labels, which are more difficult to obtain than target class labels.
Another challenge is potential performance degradation compared to backbone black-box models~\cite{Zarlenga_NeurIPS22_concept_embedding,Moayeri_ICML23_text-to-concept,Xu_ICLR24_energy-based_CBMs} due to the difficulty of learning long-tailed concept distributions~\cite{Ramaswamy_CVPR23_overlooked_factors_CBMs}.

\subsection{CBMs based on Vision-Language Models}
To address the challenges, recent works~\cite{Yuksekgonul_ICLR23_post-hoc_CBMs,Oikarinen_ICLR23_label-free_CBMs,Yang_CVPR23_LaBo} have focused on leveraging the capabilities of vision-language models (VLMs, e.g., CLIP~\cite{Radford_ICML21_CLIP}) and large language models (LLMs, e.g., GPT3~\cite{Brown_NeurIPS20_GPT3}).
These works automatically generate \(C\) in text for each \((x,y)\in\mathcal{D}\) by prompting LLM, and then, train \(g\) and \(h\) using multi-modal feature spaces with a vision encoder \(f_\mathrm{V}:\mathcal{X}\to\mathbb{R}^{d}\) and a text encoder \(f_\mathrm{T}:\mathcal{T}\to\mathbb{R}^{d}\) provided by a VLM.
We refer to such CBMs based on LLMs and VLMs as VLM-based CBMs.
As pioneering works, Post-hoc CBMs~\cite{Yuksekgonul_ICLR23_post-hoc_CBMs}, Label-free CBMs~\cite{Oikarinen_ICLR23_label-free_CBMs}, and LaBo~\cite{Yang_CVPR23_LaBo} firstly implemented this idea.
The successor works have assumed the use of LLMs or VLMs, further advancing VLM-based CBMs~\cite{Panousis_ICCV23_CDM,Rao_arXiv24_DN_CBMs,Tan_arXiv24_OpenCBM,Srivastava_arXiv_vlg_cbm}.
In particular, \cite{Panousis_ICCV23_CDM} and \cite{Rao_ECCV24_DN_CBMs} are related to our work in terms of using sparse modeling to select concepts for input images.
However, all of these existing VLM-based CBMs still require training specialized neural networks on target datasets, incurring additional training data and resources.

Handling the bi-level prediction in a zero-shot manner for unseen input is challenging because it can not be solved by na\"ive application of the existing zero-shot classification methods, which depend on limited vocabularies such as concepts related to ImageNet class names~\cite{Norouzi_ICLR14_ConSe}.
Furthermore, current VLM-based CBMs and their variants~\cite{Bhalla_NeurIPS24_splice} limit the number of concepts to a few thousand due to training and computational constraints, thereby restricting their generality.

In contrast to the previous VLM-based CBMs, the main purpose of this paper is to achieve fully zero-shot CBMs, which perform inference for input images from various domains without any additional training on target datasets.

\section{Zero-shot Concept Bottleneck Models}
\begin{figure*}[t]
\vspace{-3mm}
  \centering
  \begin{minipage}[b]{0.40\linewidth}
    \centering
    \includegraphics[width=0.9\linewidth]{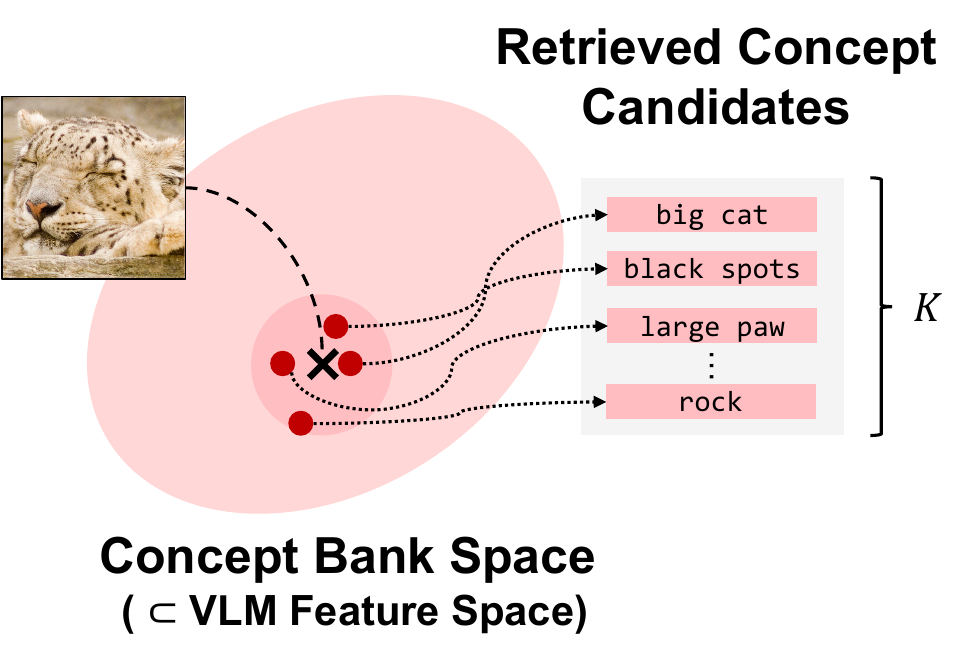}
    \subcaption{Concept Retrieval}\label{fig:concept_ret}
  \end{minipage}
  \hfill
  \begin{minipage}[b]{0.55\linewidth}
    \centering
    \includegraphics[width=0.9\linewidth]{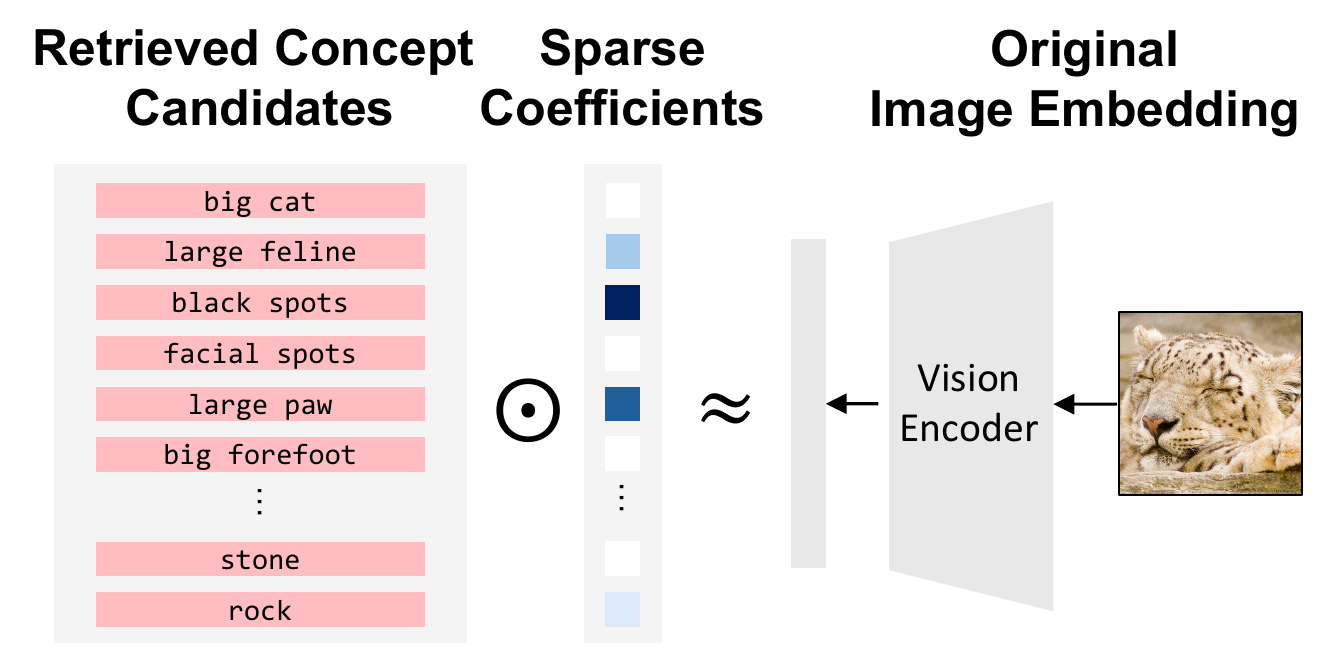}
    \subcaption{Concept Regression}\label{fig:concept_reg}
  \end{minipage}
  \vspace{-2mm}
  \caption{Concept retrieval and concept regression.
  (a) Concept retrieval searches concept candidates close to an input image in the VLM feature space and returns the top-\(K\) concepts, enabling Z-CBMs to use a large-scale concept bank for general input images.
  (b) Concept regression selects the important concepts through sparse linear regression, which approximates the input feature vectors by the weighted sum of concept candidate vectors with sparse coefficients. This sparse linear regression is helpful in selecting unique concepts.
  }
\vspace{-3mm}
\end{figure*}

In this section, we formalize the framework of Z-CBMs, which perform a zero-shot inference of input-to-concept and concept-to-class without target datasets and additional training (Fig.~\ref{fig:top}).
Z-CBMs are composed of \textit{concept retrieval} and \textit{concept regression}.
Concept retrieval finds a set of the most input-related concept candidates from millions of concepts by querying an input image feature with a semantic similarity search (Fig.~\ref{fig:concept_ret}).
Concept regression estimates the importance scores of the concept candidates by sparse linear regression to reconstruct the input feature (Fig.~\ref{fig:concept_reg}).
Finally, Z-CBMs provide the final label predicted by the reconstructed vector and concept explanations with importance scores.\looseness-1

\subsection{Zero-shot Inference Algorithm}
\textbf{Concept Retrieval.}
We first find the most semantically related concept candidates to input images from the large spaces in a concept bank \(\mathbf{C}\) (Fig.~\ref{fig:concept_ret}).
Given an input \(x\), we retrieve the set of $K$ concept candidates \(C_x \subset \mathbf{C}\) by using image and text encoders of pre-trained VLMs \(f_\mathrm{V}\) and \(f_\mathrm{T}\) as
\begin{equation}\label{eq:retrieval}
    C_x = \underset{c\in \mathbf{C}}{\operatorname{Top-K}}~\operatorname{Sim}(f_\mathrm{V}(x), f_\mathrm{T}(c)),
\end{equation}
where \({\operatorname{Top-K}}\) is an operator yielding top-\(K\) concepts in \(\mathbf{C}\) from a list sorted in descending order according to a similarity metric \(\operatorname{Sim}\).
Throughout this paper, we use cosine similarity as \(\operatorname{Sim}\) with normalized inputs by following~\cite{Conti_NeurIPS23_vocabulary_free_classification}.
Thanks to the scalability of the similarity search algorithm~\cite{Johnson_IEEE19_faiss_gpu}, Eq.~(\ref{eq:retrieval}) can efficiently find the concept candidates in an arbitrary concept bank \(\mathbf{C}\), which contains millions of concepts to describe inputs in various domains.\looseness-1

\textbf{Concept Regression.}
Given a concept candidate set \(C_x=\{c_1,...,c_K\}\), we predict the final label \(\hat{y}\) by selecting essential concepts from \(C_x\).
Conventional CBMs infer the \(C_x\)-to-\(\hat{y}\) mapping by training neural regression parameters on target tasks, which incurs the requirements of target dataset collections and additional training.
Instead, we solve this task with a different approach, leveraging the zero-shot performance of VLMs.
As shown in the previous studies~\cite{Radford_ICML21_CLIP,Jia_ICML21_ALIGN}, VLMs can be applied to zero-shot classification by inferring a label \(\hat{y}\) by matching input \(x\) and a class name text \(t_y\in \mathcal{T}\) in the multi-modal feature spaces as follows.
\begin{equation}\label{eq:zero_shot_cls}
    \hat{y} = \argmax_{y\in\mathcal{Y}}~\operatorname{Sim}(f_\mathrm{V}(x), f_\mathrm{T}(t_y)).
\end{equation}
If the feature vector \(f_\mathrm{V}(x)\) can be approximated by \(C_x\), we can achieve the zero-shot performance of black-box features by interpretable concept features.
Based on this idea, we approximate \(f_\mathrm{V}(x)\) by the weighted sum of the concept features \(F_{C_x} = [f_\mathrm{T}(c_1),...,f_\mathrm{T}(c_K)]\in\mathbb{R}^{d\times K}\) with an importance weight \(W\in\mathbb{R}^K\) (Fig.~\ref{fig:concept_reg}).
To obtain \(W\), we solve the linear regression problem defined by
\begin{equation}\label{eq:regression_obj}
\min_W \|f_\mathrm{V}(x) - F_{C_x}W\|^2_2 + \lambda\|W\|_1.
\end{equation}
Through this objective, we can achieve \(W\) not only for approximating image features but also for effectively estimating the contribution of each concept to the label prediction owing to the sparse regularization \(\|W\|_1\).
Since \(C_x\) is retrieved from a large-scale concept bank \(\mathbf{C}\), it often contains noisy concepts that are similar to each other, undermining interpretability due to semantic duplication.
In this context, the sparse regularization enhances interpretability by penalizing and eliminating unimportant concepts for the label prediction~\cite{Hastie_2015_statistical_sparsity}.

\begin{figure}[t]
\vspace{-5mm}
\begin{algorithm}[H]
    \caption{Zero-shot Inference of Z-CBMs}\label{alg:zcbm_inference}
    \begin{algorithmic}[1]
    {\footnotesize
        \REQUIRE{Input \(x\), concept bank \(C\), image encoder \(f_\mathrm{V}\), text encoder \(f_\mathrm{T}\)}
        \ENSURE{Predicted label \(\hat{y}\), concepts \(C_x\), importance weight \(W_{C_x}\)}
        \STATE{\texttt{\color{gray}\# Retrieving top-K concepts from input}}
        \STATE{\(C_x \leftarrow \retop{c}{C}{K}(f_\mathrm{V}(x), f_\mathrm{T}(c))\)}
        \STATE{\(F_{C_x} \leftarrow [f_\mathrm{T}(c_1), ..., f_\mathrm{T}(c_K)]\)}
        \STATE{\texttt{\color{gray}\# Predicting importance weights by sparse linear regression}}
        \STATE{\(W_{C_x} \leftarrow \argmin_{W\in\mathbb{R}^K} \|f_\mathrm{V}(x) - F_{C_x}W\|^2_2 + \lambda\|W\|_1\)}
        \STATE{\texttt{\color{gray}\# Predicting label by importance weighted sum concept vectors}}
        \STATE{\(\hat{y} \leftarrow \argmax_{y\in\mathcal{Y}}~\operatorname{Sim}(F_{C_x}W_{C_x}, f_\mathrm{T}(t_y))\)}
    }
    \end{algorithmic}
\end{algorithm}
\vspace{-5mm}
\end{figure}

\textbf{Final Label Prediction.}
Finally, we compute the output label with \(F_{C_x}\) and \(W\) in the same fashion as the zero-shot classification by Eq.~(\ref{eq:zero_shot_cls}), i.e.,
\begin{equation}\label{eq:zero_shot_cls_zcbms}
    \hat{y} = \argmax_{y\in\mathcal{Y}}~\operatorname{Sim}(F_{C_x}W, f_\mathrm{T}(t_y)).
\end{equation}
Algorithm~\ref{alg:zcbm_inference} shows the overall protocol of the zero-shot inference of Z-CBM.
This zero-shot inference algorithm can be applied not only to pre-trained VLMs but also to their linear probing, i.e., fine-tuning a linear head layer on the fixed feature extractor of VLMs for target tasks.

\begin{table}[t]
  \centering
  \caption{Concept Accuracy on Bird~\cite{Welinder_10_cub2002011}. Z-CBMs can infer large parts of ground-truth concepts without additional training.}
  \begin{tabular}{lc}\toprule
    Model & Accuracy \\\midrule
    CBM~\cite{Koh_ICML20_concept_bottleneck} & 71.61\\
    CDM~\cite{Panousis_ICCV23_CDM} & 45.61\\
    Z-CBM (Ours) & 60.49 \\\bottomrule
  \end{tabular}
  \label{tab:zcbm_feasibility}
  \vspace{-5mm}
\end{table}
\subsection{Feasibility Study}\label{sec:feasibility}
We show a preliminary experiment evaluating how much Algorithm~\ref{alg:zcbm_inference} can accurately infer the ground-truth concepts, given the concept set from a fully annotated dataset as the concept bank \(\mathbf{C}\).
To this end, we tested CBM~\cite{Koh_ICML20_concept_bottleneck}, CDM~\cite{Panousis_ICCV23_CDM} as a VLM-based CBM, and Z-CBM on the Bird dataset~\cite{Welinder_10_cub2002011}, which has human-annotated concept labels.
We used CLIP ViT-B/32~\cite{Radford_ICML21_CLIP} as the backbone and the annotated 312 concepts as the concept bank \(\mathbf{C}\) for CDM and Z-CBM.
More detailed protocols are described in Appendix~A.
Table~\ref{tab:zcbm_feasibility} shows the concept accuracy.
Z-CBM outperformed CDM, which requires additional training, and achieved approximately 80\% of the performance of CBM trained with the ground truth concepts.
This indicates that Z-CBMs can find valid concepts without additional training by concept retrieval and regression.
In practice, since full concept annotations are not generally given for unseen inputs, Z-CBMs adopt a large-scale concept bank that covers broad concepts for various domains.

\section{Implementation}\label{sec:implementation}
In this section, we present the detailed implementations of Z-CBMs, including backbone VLMs, concept bank construction, concept retrieval, and concept regression.

\textbf{Vision-Language Models.}
Z-CBMs allow leveraging arbitrary pre-trained VLMs for \(f_\mathrm{V}\) and \(f_\mathrm{T}\).
We basically use the official implementation of OpenAI CLIP~\citep{Radford_ICML21_CLIP} and the publicly available pre-trained weights.\footnote{https://github.com/openai/CLIP}
Specifically, by default, we use ViT-B/32 as \(f_\mathrm{V}\) and the base transformer with 63M parameters as \(f_\mathrm{T}\) by following the original CLIP.
In Section~\ref{sec:ex_multi_vlms}, we show that other VLM backbones (e.g., SigLIP~\citep{Zhai_ICCV23_SigLIP} and OpenCLIP~\citep{Cherti_CVPR23_openclip}) are also available for Z-CBMs.

\textbf{Concept Bank Construction.}
Here, we introduce the construction protocols of the concept bank \(C\) of Z-CBMs.
Since Z-CBMs can not know concepts of input image features in advance, a concept bank should contain sufficient vocabulary to describe the various domain inputs.
To this end, we extract concepts from multiple image caption datasets and integrate them into a single concept bank.
Specifically, we automatically collect concepts as noun phrases by parsing each sentence in the caption datasets, including Flickr-30K~\citep{Young_2014_Flikr30K}, CC-3M~\citep{Sharma_ACL18_CC3M}, CC-12M~\citep{Changpinyo_CVPR21_CC12M}, and YFCC-15M~\citep{Thomee_ACM16_yfcc100m}; we use the parser implemented in \texttt{nltk}~\citep{Bird_ACL06_nltk}.
At this time, the concept set size is \(|C|\approx\) 20M.
Then, we filter out nonessential concepts from the large base concept set according to policies based on \citet{Oikarinen_ICLR23_label-free_CBMs}; please see Appendix~\ref{sec:append_filtering}.
Finally, after filtering concepts, we obtain the concept bank containing \(|C|\approx\) 5M concepts. We also discuss the effect of varying caption datasets used for collecting concepts in Sec.~\ref{sec:ex_multi_dataset}~and~\ref{sec:ex_concept_bank}.

\textbf{Similarity Search in Concept Retrieval.}
Concept retrieval searches the concept candidates from input feature vectors.
To this end, we implement the concept search component by the open source library of Faiss~\citep{Johnson_IEEE19_faiss_gpu,Douze_arXiv24_faiss}.
First, we create a search index based on the text feature vectors of all concepts in a concept bank \(C\) using \(f_\mathrm{T}\).
At inference time, we retrieve the concept vectors via similarity search on the concept index by specifying the concept number \(K\).
We set \(K=2048\) as the default value and empirically show the effect of \(K\) in Appendix~\ref{sec:ex_concept_search}.\looseness-1

\textbf{Sparse Linear Regression in Concept Regression.}
In concept regression, we can use arbitrary sparse linear regression algorithms, including lasso~\citep{Tibshirani_1996_lasso}, elastic net~\citep{Zou_2005_elasticnet}, and sparsity-constrained optimization like hard thresholding pursuit~\citep{Yuan_ICML14_sparsity-constrained_htp}.
The efficient implementations of these algorithms are publicly available on the {\tt sklearn}~\citep{sklearn} and {\tt skscope}~\citep{skscope} libraries.
The choice of sparse linear regression algorithm depends on the use cases. 
For example, lasso is useful when one wants to naturally obtain important concepts from a large number of candidate concepts, elastic net is effective for high target task performance, and sparsity-constrained optimization satisfies rigorous requirements regarding the number of concepts for explanations.
We use lasso with \(\lambda=1.0\times10^{-5}\) as the default algorithm (see Appendix~\ref{sec:append_setting}~and~\ref{sec:ex_lambda}), but we confirm that arbitrary sparse linear regression algorithms are available for Z-CBMs in Sec~\ref{sec:ex_analysis}.

\section{Experiments}\label{sec:experiment}
We evaluate Z-CBMs on multiple visual classification datasets and pre-training VLMs.
We conduct experiments on two scenarios: \textit{zero-shot} and \textit{training head}; the former uses pre-trained VLMs for inference without any training, while the latter learns only the classification heads.

\subsection{Settings}\label{sec:ex_settings}
\textbf{Datasets.}
By following previous studies~\cite{Radford_ICML21_CLIP}, we evaluated Z-CBMs on 12 diverse image classification datasets: \textbf{Aircraft (Air)}~\cite{maji_13_aircraft}, \textbf{Bird}~\cite{Welinder_10_cub2002011}, \textbf{Caltech-101 (Cal)}~\cite{FeiFei_caltech101} \textbf{Car}~\cite{krause_3DRR2013_stanford_cars}, \textbf{DTD}~\cite{cimpoi_CVPR14_DTD}, \textbf{EuroSAT (Euro)}~\cite{Helber_IEEE_eurosat}, \textbf{Flower (Flo)}~\cite{Nilsback_08_flowers}, \textbf{Food}~\cite{bossard14_Food101}, \textbf{ImageNet (IN)}~\cite{russakovsky_imagenet}, \textbf{Pet}~\cite{parkhi_CVPR12_oxford_pets}, \textbf{SUN397}~\cite{Xiao_CVPR10_sun397}, and \textbf{UCF-101}~\cite{Soomro_arXiv12ucf101}.
In the training head scenario, we randomly split a training dataset into \(9:1\) and used the former as the training set and the latter as the validation set.
For ImageNet, we set the split ratio \(99:1\).

\textbf{Zero-shot Baselines.}
For the zero-shot baseline, our Z-CBMs with the zero-shot inference of a black-box VLM and ConSe~\cite{Norouzi_ICLR14_ConSe}, which is a zero-shot classification method predicting a class label with a weighted sum of ImageNet concept features.
For more details, please see Appendix~\ref{sec:append_setting}.

\textbf{Training Head Baselines.}
To compare Z-CBMs with existing VLM-based CBMs, we evaluated models trained on target datasets.
In this setting, Z-CBMs were applied to linear probing of VLMs, i.e., fine-tuning only a linear head layer on the feature extractors of VLMs; we refer to this pattern LP-Z-CBM.
As the baselines, we used \textbf{Lable-free CBM}~\cite{Oikarinen_ICLR23_label-free_CBMs}, \textbf{LaBo}~\cite{Yang_CVPR23_LaBo}, and \textbf{CDM}~\cite{Panousis_ICCV23_CDM}.
We performed these methods based on their repositories.\looseness-1

\textbf{Evaluation Metrics.}
To evaluate predicted concepts, we used the \textbf{SigLIP-Score}, the cosine similarity between image and text embeddings on SigLIP~\cite{Zhai_ICCV23_SigLIP} (higher is better).
This score indicates how well a predicted concept explains an image~\cite{Radford_ICML21_CLIP,Hessel_EMNLP21_clipscore}, serving as a quality indicator for input-to-concept inference. 
Specifically, we averaged SigLIP-Scores between test images and their predicted concept texts for the top 10 concepts, ranked by absolute importance scores.
Concretely, we measured the average SigLIP-Scores between test images and the predicted concept texts, where we extracted the top 10 concepts from sorted concepts in descending order by absolute concept importance scores for each model.
We also used \textbf{concept recall} to evaluate Z-CBM's predicted concepts.
Top-\(K\) concept recall \({|C^\mathrm{Z} \cap C^\mathrm{R}|}/{K}\) measures the overlap between Z-CBM's top-\(K\) concepts \(C^\mathrm{Z}=\{c^\mathrm{Z}_i\}^K_{i=1}\subset \mathbf{C}\) (with non-zero coefficients) and \(N_\mathrm{R}\) reference concepts \(C^\mathrm{R}=\{c^\mathrm{R}_i\}^{N_\mathrm{R}}_{i=1}\subset \mathbf{C}\) from VLM-based CBMs requiring training.
This metric assesses concept overlap between Z-CBMs and a training-based CBM using the same concept bank \(\mathbf{C}\), indicating Z-CBM's approximation of a trained model's concepts. 
Specifically, we averaged concept recall at \(K=10\) across test samples, using the GPT-generated concept banks~\cite{Oikarinen_ICLR23_label-free_CBMs}, and reference concepts of Label-free CBMs
Following~\cite{Oikarinen_ICLR23_label-free_CBMs}, \(C^\mathrm{R}\) comprised concepts with contribution scores $>0.05$.
Finally, we report \textbf{top-1 test accuracy} for target classification performance.~\looseness-1

\subsection{Quantitative Evaluation of Predicted Concepts}\label{sec:ex_quantitative_eval_concept}

We first quantitatively evaluate the predicted concepts of Z-CBMs from the perspective of their factuality to represent image features.
We measured averaged CLIP-Score and concept coverage across the 12 datasets.

Table~\ref{tb:clip_score} shows the results of CLIP-Score. 
For all datasets, our Z-CBM predicted concepts that are strongly correlated to input images, and it largely outperformed the CBM baselines that require training.
This can be caused by the choice of concept bank.
Existing CBMs perform concept-to-label inference with learnable parameters, making it difficult to handle millions of concepts at once.
Thus, they often limit their concept vocabularies to a few thousand to ensure learnability.
In contrast, our Z-CBMs can treat millions of concepts without training by dynamically retrieving concepts of interest and inferring essential concepts with sparse linear regression.
That is, paradoxically, Z-CBMs succeed in providing accurate image explanations through an abundant concept vocabulary by eliminating training.

On the other hand, Table~\ref{tb:concept_coverage} shows the results of concept coverage when using the concepts predicted by Label-free CBMs as the reference concepts. 
We also list the results of Z-CBMs using cosine similarity on CLIP and linear regression to compute the importance coefficients instead of lasso; since all of their coefficients are non-zero values, we measured the concept coverage scores by using the top 128 concepts.
Z-CBMs with lasso achieved the best concept coverage; the average score was 85.27\%.
This indicates that Z-CBMs can predict most of the important concepts found by trained CBMs, and sparse linear regression is a key factor for finding important concepts without training.

\begin{table}[!tbp]
  \begin{minipage}[h]{0.4\linewidth}
  \centering
  \caption{
    SigLIP-Score
   }
  \label{tb:clip_score}%
  \resizebox{\columnwidth}{!}{
    \begin{tabular}{lc}
    \toprule
    \multicolumn{1}{l}{\textbf{Method}} & \textbf{Avg. of 12 datasets} \\
    \midrule
    Label-free CBM & 0.5485  \\
    LaBo & 0.5419\\
    CDM & 0.5714 \\
    \rowcolor{blue!20}  
    Z-CBM (ALL) & \textbf{0.6309} \\
    \bottomrule
    \end{tabular}
    } 
  \end{minipage}
  \hfill
  \begin{minipage}[h]{0.58\linewidth}
  \centering
  \caption{
    Concept Recall (\%)
   }
  \label{tb:concept_coverage}%
  \resizebox{\textwidth}{!}{
    \begin{tabular}{lc}
    \toprule
    \multicolumn{1}{l}{\textbf{Method}} & \textbf{Avg. of 12 datasets} \\
    \midrule
    Z-CBM (Cosine Similarity) & 58.51\\  
    Z-CBM (Linear Regression) & 76.87 \\
    \rowcolor{blue!20}  
    Z-CBM (Lasso) & \textbf{85.27} \\
    \bottomrule
    \end{tabular}
    }
  \end{minipage}
  \vspace{-3mm}
\end{table}%

\subsection{Evaluation of Human Intervention}\label{sec:ex_reliability}

\begin{figure}[t]
  \centering
  \begin{minipage}[h]{0.48\linewidth}
    \centering
    \includegraphics[width=1.0\linewidth]{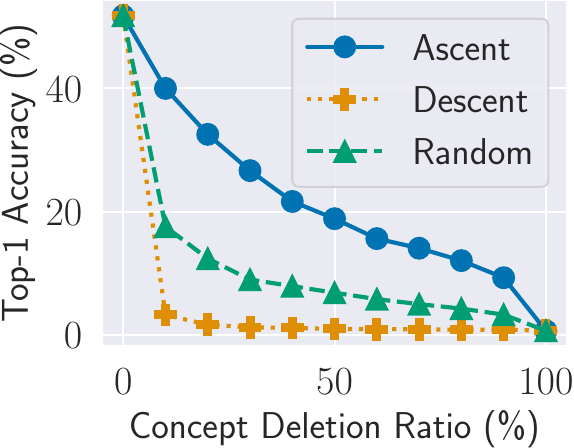}
    \vspace{-2mm}
    \caption{Concept Deletion}\label{fig:concept_deletion}
  \end{minipage}
  \hfill
  \begin{minipage}[h]{0.48\linewidth}
    \centering
    \includegraphics[width=1.0\linewidth]{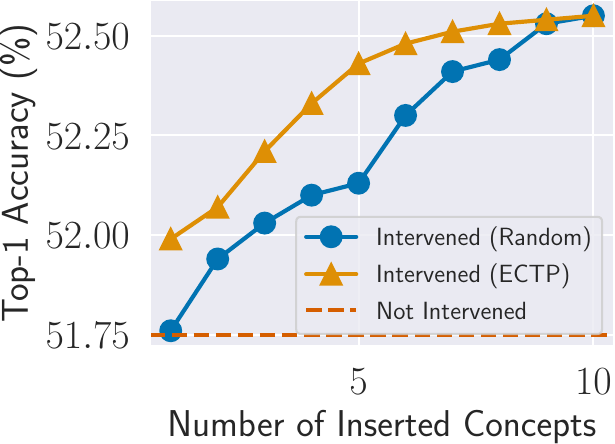}
    \vspace{-2mm}
    \caption{Concept Insertion}\label{fig:concept_insertion}
  \end{minipage}
\vspace{-5mm}
\end{figure}

Human intervention in concepts is an essential feature shared by the CBM family for debugging models and modifying the output to make the final prediction accurate.
In addition to interventions in existing concepts in the concept bank, Z-CBMs allows interventions in arbitrary concepts described in natural language.
We evaluate Z-CBMs via two types of intervention: (i) concept deletion and (ii) concept insertion.
In concept deletion, we confirm the dependence on the predicted concepts by removing the concept with non-zero coefficients in ascending, descending, and random orders.
Fig.~\ref{fig:concept_deletion} shows the results on Bird by varying the deletion ratio.
The accuracy of Z-CBMs largely dropped with the smaller deletion ratio in the descent cases, indicating that Z-CBM selects the important concepts via concept regression and relies on them for the final prediction.
In the ascent cases, the accuracy slowly and steadily decreases, suggesting that the Z-CBMs are not biased toward limited concepts and that all of the selected concepts are essential.

In concept insertion, we first predict concepts by concept regression and add randomly selected ground-truth concepts to the output non-zero concept set.
Then, we re-run concept regression with linear regression on this modified concept set and predict the final label prediction by Eq.~(\ref{eq:zero_shot_cls_zcbms}).
As the ground truth concepts, we used the attribute labels of Bird~\cite{Welinder_10_cub2002011}.
Fig.~\ref{fig:concept_insertion} shows the top-1 accuracy of the intervened Z-CBMs.
In addition to random selection, we performed a sophisticated intervention method called ECTP~\cite{Shin_ICML23_intervention_cbm}.
The performance improved as the number of inserted concepts per sample increased for both cases.
This indicates that Z-CBMs can correct the final output by modifying the concept of interest through intervention.

\subsection{Qualitative Evaluation of Predicted Concepts}\label{sec:ex_qualitative_eval_concept}
\begin{figure*}[t]
  \centering
    \centering
    \includegraphics[width=\linewidth]{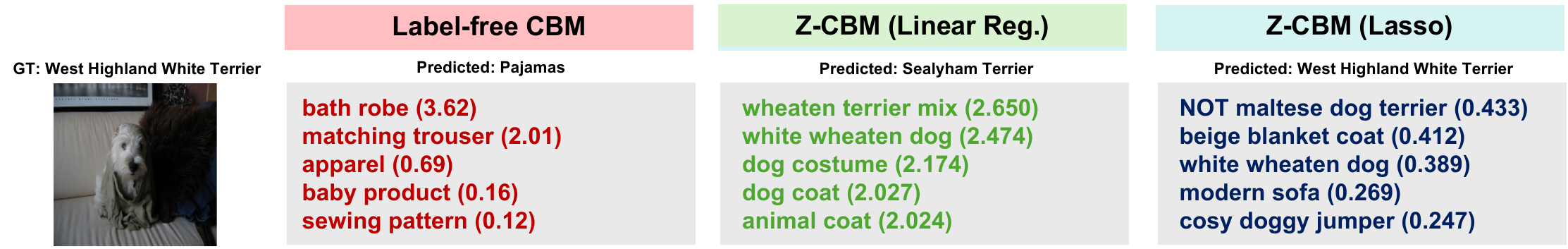}
    \caption{
    Qualitative evaluation of predicted concepts on the ImageNet validation set.
    While Label-free CBMs sometimes hallucinate invisible concepts or ignore important concepts, Z-CBMs with lasso provide realistic and dominant concepts in inputs with diverse vocabulary. \textbf{NOT} prefix denotes that the concept has negative coefficients similar to~\cite{Oikarinen_ICLR23_label-free_CBMs}~and~\cite{Panousis_ICCV23_CDM}.
    }\label{fig:qualitative_eval}
  \vspace{-3mm}
\end{figure*}

We demonstrate the qualitative evaluation of predicted concepts by Label-free CBMs, Z-CBMs with Liner Reg$.$, and Z-CBMs with Lasso when inputting the ImageNet validation examples in Fig.~\ref{fig:qualitative_eval}.
Overall, Z-CBMs tend to accurately predict realistic and dominant concepts that appear in input images, even though they are not trained on target tasks.
For instance, Z-CBM predicts various concepts related to dogs, clothes, and background, whereas Label-free CBM focuses on clothes and ignores dogs and background.
This difference may be caused by the fact that the image-to-concept mapping of Z-CBMs is not biased toward the label information because it does not train on the target data.
For the comparison of linear regression and lasso,  Z-CBM (Linear Reg$.$) tends to produce semantically overlapped concepts.
In fact, quantitatively, we also found that the averaged inner SigLIP-Scores among the top-10 concepts of lasso (0.5552) is significantly lower than that of linear regression (0.7425).
These results emphasize the advantage of using sparse modeling in concept regression to select mutually exclusive concepts from the large concept bank.\looseness-1

\subsection{Zero-shot Image Classification Performance}\label{sec:ex_multi_dataset}
Table~\ref{tb:multiple_dataset} summarizes the averaged top-1 accuracy across the 12 image classification datasets.
It also shows the ablation of concept banks; the brackets in the Z-CBM rows represent the dataset used to construct the concept bank.
In the zero-shot setting, we observed that our Z-CBMs outperformed the zero-shot CLIP baseline.
This is beyond our expectations and may be due to the fact that Z-CBMs approximate image features with the weighted sum of concept text features, reducing the modality gap between the original image and the label text (see Appendix~\ref{sec:ex_modality_gap}).
The ablation of concept banks demonstrates that higher accuracy tends to be achieved by larger concept banks.
This indicates that image features are more accurately approximated by selecting concepts from a rich vocabulary.
We further explore the impacts of concept banks in Sec.~\ref{sec:ex_concept_bank}.

In the training head setting, Z-CBMs based on linear probing models (\textbf{LP-Z-CBMs}) reproduced the accuracy of linear probing well.
Further, LP-Z-CBMs stably outperformed existing methods that require additional training for special modules.
This suggests that our concept retrieval and concept regression using the original CLIP features are sufficient for input-to-concept and concept-to-label inference in terms of target task performance.

\subsection{Detailed Analysis}\label{sec:ex_analysis}

\subsubsection{Effects of Backbone VLMs}\label{sec:ex_multi_vlms}
We show the impacts on Z-CBMs when varying backbone VLMs.
Since vision-language models are being intensively studied, it is important to confirm the compatibility of Z-CBMs with successor models with better zero-shot performance.
In addition to the CLIP models, we used OpenCLIP~\citep{Cherti_CVPR23_openclip}, SigLIP~\citep{Zhai_ICCV23_SigLIP}, and DFN~\citep{Fang_ICLR24_DFN}.
Table~\ref{tb:vlm_backbone} demonstrates the results, including the original zero-shot classification accuracy and the accuracy with Z-CBMs, and CLIP-Score.
The performance of Z-CBMs improved in proportion to the zero-shot performance of the VLMs.
In particular, the gradual improvement in CLIP-Score indicates that input-to-concept inference becomes more accurate with more powerful VLMs.
We also observed that the improvement phenomenon over black-box baselines discussed in Sec.~\ref{sec:ex_multi_dataset} appears especially in small models where the multi-modal alignment capability is relatively weak.
These results suggest that Z-CBM is universally applicable across generations of VLMs, and that its practicality will improve as VLMs evolve in future work.

\subsubsection{Effects of Concept Bank}\label{sec:ex_concept_bank}
As shown in Sec.~\ref{sec:ex_multi_dataset} and Table~\ref{tb:multiple_dataset}, the choice of concept bank is crucial for the performance.
Here, we provide a more detailed analysis of the concept banks.
Table~\ref{tb:concept_bank} summarizes the results when varying concept banks.
For comparison, we added the concept bank generated by GPT-3 from ImageNet class names, which is used in Label-free CBMs~\citep{Oikarinen_ICLR23_label-free_CBMs}; we used the concept sets published in the official repository.
Although it is competitive with the existing CBM baseline (Label-free CBMs), Z-CBMs with the GPT-3 concepts significantly degraded the top-1 accuracy from Zero-shot CLIP, and the CLIP score was much lower than that of our concept banks composed of noun phrases extracted from caption datasets.
This indicates that the concept bank used in the existing method is limited in its ability to represent image concepts.
Meanwhile, our concept bank scalably improved in accuracy and CLIP-Score as its size increased, and combining all of them achieved the best results. \looseness-1

\setlength{\tabcolsep}{6pt}
\begin{table}[!tbp]
  \centering
  \caption{
    Top-1 accuracy on 12 classification datasets with CLIP ViT-B/32.
    Complete results appear in Appendix~D-A.
   }
  \label{tb:multiple_dataset}
  \resizebox{0.8\columnwidth}{!}{
    \begin{tabular}{llc}
    \toprule
    \multicolumn{1}{l}{\textbf{Setting}} &\multicolumn{1}{l}{\textbf{Method}} & \textbf{Avg. of 12 datasets} \\
    \midrule
    \multirow{6}{*}{Zero-Shot} & \multicolumn{1}{l}{Zero-shot CLIP} & 53.73  \\
                               & \multicolumn{1}{l}{ConSe} & 10.82  \\
    \cmidrule{2-3}
    & \multicolumn{1}{l}{Z-CBM (Flickr30K)} & 52.62\\
    & \multicolumn{1}{l}{Z-CBM (CC3M)} & 52.98 \\
    & \multicolumn{1}{l}{Z-CBM (CC12M)} & 53.97 \\
    & \multicolumn{1}{l}{Z-CBM (YFCC15M)} & 53.94  \\
    \rowcolor{blue!20}  
    \cellcolor{white} & \multicolumn{1}{l}{Z-CBM (ALL)} & \textbf{54.28}\\
    \midrule
    \multirow{5}{*}{Training Head} & \multicolumn{1}{l}{Linear Probe CLIP} & \textbf{78.98} \\\cmidrule{2-3}
    & \multicolumn{1}{l}{Label-free CBM} & 74.87 \\
    & \multicolumn{1}{l}{LaBo} & 74.04 \\
    & \multicolumn{1}{l}{CDM} & 76.39\\
    \rowcolor{blue!20}  
    \cellcolor{white} & \multicolumn{1}{l}{LP-Z-CBM (ALL)} & \text{78.31}\\
    \bottomrule
    \end{tabular}
    }
\end{table}%

\begin{table}[!tbp]
  \centering
  \caption{
    Performance of Z-CBMs varying backbone VLMs on ImageNet.
   }
  \label{tb:vlm_backbone}%
  \vspace{-3mm}
  \resizebox{1.0\columnwidth}{!}{
    \begin{tabular}{lccc}
    \toprule
    \multirow{2}{*}{\textbf{Backbone VLM}} & \textbf{Top-1 Acc.} & \textbf{Top-1 Acc.}  & \textbf{SigLIP-Score}  \\
    & \textbf{(Black Box)} & \textbf{(Z-CBM)}  & \textbf{(Z-CBM)} \\
    \midrule
    CLIP ViT-B/32        & 61.88 & 62.70 & 0.6498 \\  
    CLIP ViT-L/14        & 72.87 & 73.19 & 0.6608 \\
    OpenCLIP ViT-H/14    & 77.20 & 77.81 & 0.6790 \\
    OpenCLIP ViT-G/14    & 79.03 & 78.27 & 0.6810 \\
    DFN ViT-H/14         & 83.85 & 83.40 & 0.7038 \\
    \bottomrule
    \end{tabular}
    }
  \vspace{-3mm}
\end{table}%

\begin{table}[!tbp]
  \centering
  \caption{
    Performance of Z-CBMs varying concept banks on ImageNet with CLIP ViT-B/32.
   }
  \label{tb:concept_bank}%
  \vspace{-3mm}
  \resizebox{1.0\columnwidth}{!}{
    \begin{tabular}{lccc}
    \toprule
    \multicolumn{1}{l}{\textbf{Concept Bank}} & \textbf{Vocab. Size} & \textbf{Top-1 Acc.} & \textbf{SigLIP-Score} \\
    \midrule
    Zero-shot CLIP & N/A & 61.88 & N/A \\
    \midrule
    Label-free CBM w/ GPT-3 (ImageNet Class) & 4K& 58.00 & 0.5896 \\
    CDM w/ GPT-3 (ImageNet Class) & 4K & 62.52 & 0.6193 \\
    \midrule
    GPT-3 (ImageNet Class) & 4K & 59.18 & 0.5407  \\  
    Noun Phrase (Flickr30K) & 45K & 61.52 & 0.5539 \\
    Noun Phrase (CC3M) & 186K & 62.38 & 0.5904 \\
    Noun Phrase (CC12M) & 2.58M & 62.42 & 0.6242  \\
    Noun Phrase (YFCC15M) & 2.20M & 62.45 & 0.6375  \\
    \rowcolor{blue!20}  
    Noun Phrase (ALL) & \textbf{5.12M} & \textbf{62.70} & \textbf{0.6498} \\
    \bottomrule
    \end{tabular}
    }
  \vspace{-3mm}
\end{table}%

\section{Conclusion}
In this paper, we presented zero-shot CBMs (Z-CBMs), which predict input-to-concept and concept-to-label mappings in a fully zero-shot manner.
To this end, Z-CBMs first search input-related concept candidates by concept retrieval, which leverages pre-trained VLMs and a large-scale concept bank containing millions of concepts to explain outputs for unseen input images in various domains.
For the concept-to-label inference, concept regression estimates the importance of concepts by solving the sparse linear regression approximating the input image features with linear combinations of selected concepts.
Our extensive experiments show that Z-CBMs can provide interpretable and intervenable concepts comparable to conventional CBMs that require training.
Since Z-CBMs can be built on any off-the-shelf VLMs, we believe that it will be a good baseline for zero-shot interpretable models based on VLMs in future research.


\bibliography{ref}
\bibliographystyle{icml2026}

\newpage
\appendix
\onecolumn

\section{Details of Concept Filtering}\label{sec:append_filtering}
We basically follow the policies introduced by \cite{Oikarinen_ICLR23_label-free_CBMs}, which removes (i) too long concepts, (ii) too similar concepts to each other, and (iii) too similar concepts to target class names (optional).
However, the second policy is computationally intractable because it requires the \(\mathcal{O}(|C|^2)\) computation of the similarity matrix across all concepts.
Thus, we approximate this using a similarity search by Eq.~(\ref{eq:retrieval}) that yields the most similar concepts.
We retrieve the top 64 concepts from a concept and remove them according to the original policy.

\section{Details of Settings}\label{sec:append_setting}
\paragraph{Zero-shot Baselines.}
For the black-box baseline, according to the previous work~\citep{Radford_ICML21_CLIP}, we construct a class name prompt \(t_y\) by the scheme of ``\texttt{a photo of [class name]}'', and make VLMs predict a target label \(\hat{y}\) by Eq.~(\ref{eq:zero_shot_cls}).
ConSe is a zero-shot cross-modal classification method that infers a target label from a semantic embedding composed of the weighted sum of concepts of the single predicted ImageNet label.
For Z-CBMs, we selected $1.0\times10^{-5}$ as $\lambda$ by searching from $\{1.0\times10^{-2},1.0\times10^{-3},1.0\times10^{-4},1.0\times10^{-5},1.0\times10^{-6},1.0\times10^{-7},1.0\times10^{-8}\}$ to choose the minimum value achieving over 10\% non-zero concept ration when using $K=2048$ on the subset of ImageNet training set. We used the same $\lambda$ for all experiments.
To make 

\section{Additional Experiments}
\subsection{Detailed Results for All Datasets}
Table~\ref{tb:clip_score_full}, \ref{tb:concept_coverage_full}, and~\ref{tb:multiple_dataset_full} shows all of the results on the 12 datasets omitted in Table~\ref{tb:clip_score}, \ref{tb:concept_coverage}, and~\ref{tb:multiple_dataset}, respectively.

\setlength{\tabcolsep}{6pt}
\begin{table*}[!tbp]
  \centering
  \caption{
    CLIP-Score on 12 classification datasets. 
    We compute the averaged CLIP-Scores between images and concepts with top-10 absolute coefficients.
   }
  \label{tb:clip_score_full}%
  \vspace{-3mm}
  \resizebox{1.0\textwidth}{!}{
    \begin{tabular}{lccccccccccccc}
    \toprule
    \multicolumn{1}{l}{\textbf{Method}} & \textbf{Air} & \textbf{Bird} & \textbf{Cal} & \textbf{Car} & \textbf{DTD} & \textbf{Euro} & \textbf{Flo} & \textbf{Food} & \textbf{IN} & \textbf{Pet} & \textbf{SUN} & \textbf{UCF} & \textbf{Avg.} \\
    \midrule
    Label-free CBM &0.6824 & 0.7818 & 0.7023 & 0.7106 & 0.6552 & 0.6179 & 0.6988 & 0.6959 & 0.7202 & 0.7119 & 0.7327 & 0.6688 & 0.6982  \\
    LaBo & 0.6980 & 0.7626 & 0.7211 & 0.7411 & 0.6299 & 0.6202 & 0.7138 & 0.7526 & 0.7272 & 0.7235 & 0.7060 & 0.6978 & 0.7078\\
    CDM & 0.6887 & 0.7655 & 0.7164 & 0.7221 & 0.7000 & 0.6584 & 0.7239 & 0.7151 & 0.7618 & 0.7257 & 0.7049 & 0.6870 & 0.7141 \\
    \rowcolor{blue!20}  
    Z-CBM (ALL) &\textbf{0.7811} & \textbf{0.8100} & \textbf{0.7748} & \textbf{0.7582} & \textbf{0.7661} & \textbf{0.7457} & \textbf{0.7767} & \textbf{0.7785} & \textbf{0.7766} & \textbf{0.7477} & \textbf{0.7925} & \textbf{0.7965} & \textbf{0.7754} \\
    \bottomrule
    \end{tabular}}
  \vspace{-3mm}
\end{table*}%
\begin{table*}[!tbp]
  \centering
  \caption{
    Concept coverage (\%) of Z-CBMs on 12 classification datasets}
  \label{tb:concept_coverage_full}%
  \vspace{-3mm}
  \resizebox{1.0\textwidth}{!}{
    \begin{tabular}{lccccccccccccc}
    \toprule
    \multicolumn{1}{l}{\textbf{Method}} & \textbf{Air} & \textbf{Bird} & \textbf{Cal} & \textbf{Car} & \textbf{DTD} & \textbf{Euro} & \textbf{Flo} & \textbf{Food} & \textbf{IN} & \textbf{Pet} & \textbf{SUN} & \textbf{UCF} & \textbf{Avg.} \\
    \midrule
    Z-CBM (Cosine Similarity) & 66.83 & 41.42 & 37.13 & 60.95 & 71.85 & 90.37 & 50.39 & 77.50 & 48.80 & 90.07 & 29.76 & 37.04 & 58.51\\  
    Z-CBM (Linear Regression) & 96.45 & 81.98 & 51.82 & 58.06 & 91.40 & 90.91 & 90.82 & 90.88 & 71.51 & 95.37 & 40.84 & 62.43 & 76.87 \\
    \rowcolor{blue!20}  
    Z-CBM (Lasso) & \textbf{98.95} & \textbf{86.01} & \textbf{69.97} & \textbf{96.43} & \textbf{94.26} & \textbf{91.91} & \textbf{93.57} & \textbf{96.74} & \textbf{86.92} & \textbf{97.37} & \textbf{42.86} & \textbf{68.20} & \textbf{85.27} \\
    \bottomrule
    \end{tabular}}
  \vspace{-3mm}
\end{table*}%
\begin{table*}[!tbp]
  \centering
  \caption{
    Top-1 accuracy on 12 classification datasets with CLIP ViT-B/32.
   }
  \label{tb:multiple_dataset_full}
  \vspace{-3mm}
  \resizebox{1.0\textwidth}{!}{
    \begin{tabular}{llccccccccccccc}
    \toprule
    \multicolumn{1}{l}{\textbf{Setting}} &\multicolumn{1}{l}{\textbf{Method}} & \textbf{Air} & \textbf{Bird} & \textbf{Cal} & \textbf{Car} & \textbf{DTD} & \textbf{Euro} & \textbf{Flo} & \textbf{Food} & \textbf{IN} & \textbf{Pet} & \textbf{SUN} & \textbf{UCF} & \textbf{Avg.} \\
    \midrule
    \multirow{6}{*}{Zero-Shot} & \multicolumn{1}{l}{Zero-shot CLIP} & 18.93  & 51.80  & 24.50  & 60.38  & 43.24  & 35.54  & 63.41  & 78.61  & 61.88  & 85.77  & 61.21  & 59.48 & 53.73  \\
    \cmidrule{2-15}
    & \multicolumn{1}{l}{Z-CBM (Flickr30K)} & 18.27 & 46.70 & 24.26 & 56.46 & 43.56 & 34.32 & 59.80 & 78.17 & 61.52 & 85.46 & 62.23 & 60.67 & 52.62\\
    & \multicolumn{1}{l}{Z-CBM (CC3M)} & 18.09 & 48.53 & 24.30 & 55.58 & 43.51 & 35.09 & 61.44 & 78.89 & 62.68 & 85.29 & 62.18 & 60.45 & 52.98 \\
    & \multicolumn{1}{l}{Z-CBM (CC12M)} & 18.66 & 51.03 & 24.42 & \textbf{59.22} & 43.72 & \textbf{36.73} & 63.31 & 79.26 & 62.42 & \textbf{85.98} & 62.11 & 60.75 & 52.98 \\
    & \multicolumn{1}{l}{Z-CBM (YFCC15M)} & 18.81 & \textbf{51.87} & 24.54 & 58.72 & 43.40 & 35.96 & 63.38 & 79.22 & 62.42 & 85.94 & 62.07 & 60.96 & 53.97  \\
    \rowcolor{blue!20}  
    \cellcolor{white} & \multicolumn{1}{l}{Z-CBM (ALL)} & \textbf{19.00} & 51.75 & \textbf{25.42} & 58.87 & \textbf{43.86} & 36.12 & \textbf{63.78} & \textbf{82.44} & \textbf{62.70} & 85.95 & \textbf{62.89} & \textbf{61.49} & \textbf{54.28}\\
    \midrule
    \multirow{5}{*}{Training Head} & \multicolumn{1}{l}{Linear Probe CLIP} & \textbf{45.06} & \textbf{72.72} & \textbf{95.70} & \textbf{79.75} & \textbf{74.84} & \textbf{92.99} & \textbf{94.02} & \textbf{87.06} & \textbf{68.54} & \textbf{88.72} & \textbf{65.20} & \textbf{83.14} & \textbf{78.98} \\\cmidrule{2-15}
    & \multicolumn{1}{l}{Label-free CBM} & 42.72 & 67.05 & 94.12 & 71.81 & \textbf{74.31} & 91.30 & 91.23 & 81.91 & 58.00 & 83.29 & 62.00 & 80.68 & 74.87 \\
    & \multicolumn{1}{l}{LaBo} &43.43 & 69.38 & 94.82 & 77.78 & 73.59 & 88.17 & 91.67 & 84.29 & 59.16 & 87.24 & 57.70 & 81.26 & 74.04 \\
    & \multicolumn{1}{l}{CDM} & 44.58 & 69.75 & 95.78 & 77.27 & 74.80 & \textbf{92.16} & 92.99 & 81.85 & 62.52 & 86.59 & 56.48 & 81.93 & 76.39\\
    \rowcolor{blue!20}  
    \cellcolor{white} & \multicolumn{1}{l}{LP-Z-CBM (ALL)} & \text{44.80} & \text{71.67} & \text{95.50} & \text{78.09} & 73.94 & 91.22 & \text{93.28} & \text{86.73} & \text{67.99} & \text{88.58} & \text{65.53} & \text{82.37} & \text{78.31}\\
    \bottomrule
    \end{tabular}}
  \vspace{-3mm}
\end{table*}%

\begin{figure*}[!tbp]
  \centering
\begin{minipage}[h]{0.48\linewidth}
    \centering
    \includegraphics[width=\textwidth]{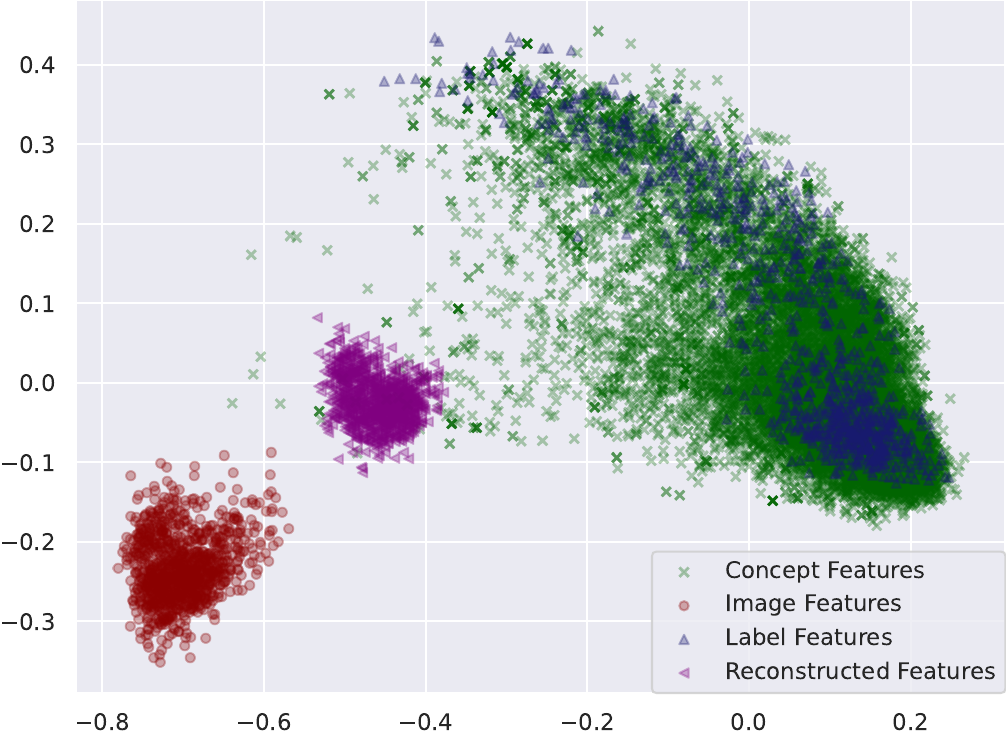}
    \caption{PCA feature visualization of Z-CBMs}\label{fig:pca_feat_viz}
  \end{minipage}
  \hfill
  \begin{minipage}[h]{0.48\linewidth}
    \centering
    \includegraphics[width=\textwidth]{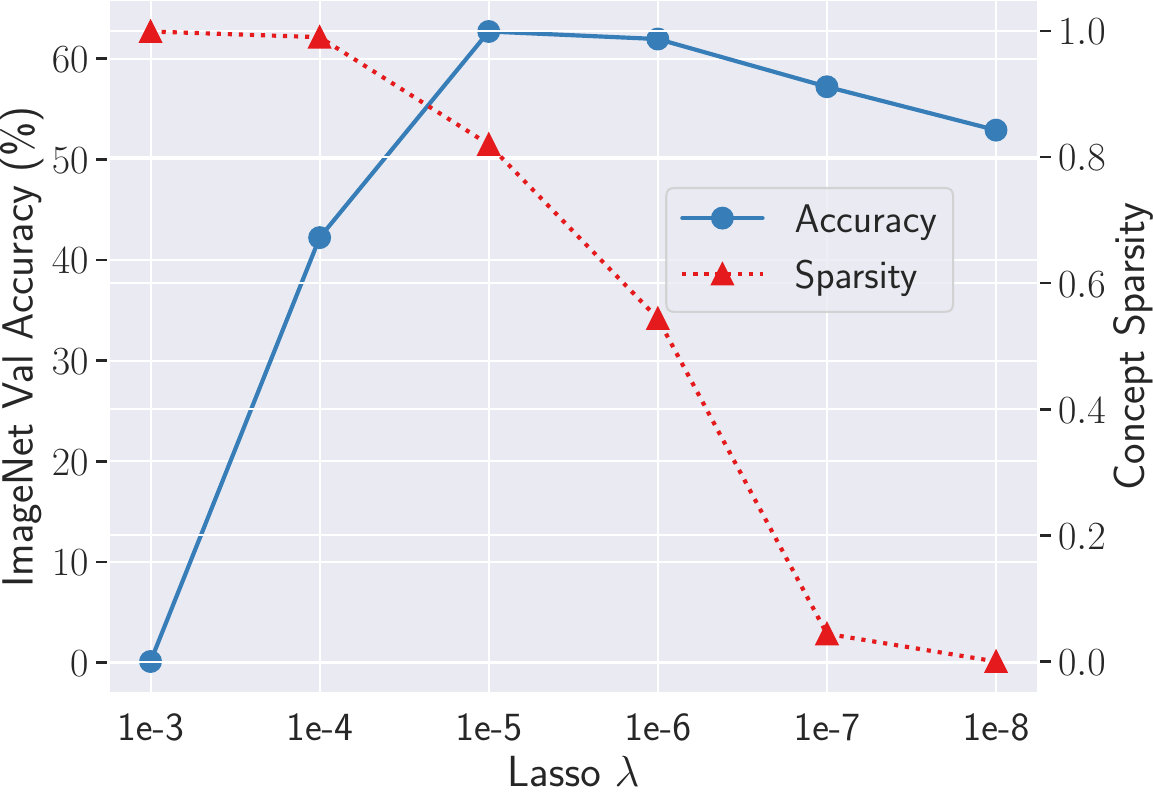}
    \caption{Effects of varying $\lambda$ in Eq.~\ref{eq:regression_obj}}\label{fig:lambda}
  \end{minipage}
  \vspace{-3mm}
\end{figure*}%

\subsection{Analysis on Modality Gap}\label{sec:ex_modality_gap}
In Section~\ref{sec:ex_multi_dataset}, Table~\ref{tb:multiple_dataset} shows that Z-CBMs improved the zero-shot CLIP baselines.
We hypothesize that the reason is reducing the modality gap~\citep{Liang_NeurIPS22_clip_modality_gap} between image and text features by the weighted sum of concept features to approximate \(f_\mathrm{V}(x)\) by Eq.~\ref{eq:regression_obj}.
To confirm this, we conduct a deeper analysis of the effects of Z-CBMs on the modality gap with quantitative and qualitative evaluations.
For quantitative evaluation, we measured the L2 distance between image-label features and concept-label features as the modality gap by following~\citep{Liang_NeurIPS22_clip_modality_gap}.
The L2 distances were $1.74\times 10^{-3}$ in image-to-label and $0.86\times 10^{-3}$ in concept-to-label, demonstrating that Z-CBMs largely reduce the modality gap by concept regression.
We also show the PCA feature visualizations in Figure~\ref{fig:pca_feat_viz}, indicating that the weighted sums of concepts (reconstructed concepts) bridge the image and text modalities.

\subsection{Effects of $\lambda$}\label{sec:ex_lambda}
Here, we discuss the effects when changing $\lambda$ in Eq.~(\ref{eq:regression_obj}).
We varied $\lambda$ in $\{1.0\times10^{-2},1.0\times10^{-3},1.0\times10^{-4},1.0\times10^{-5},1.0\times10^{-6},1.0\times10^{-7},1.0\times10^{-8}\}$.
Figure~\ref{fig:lambda} plots the accuracy and the sparsity of predicted concepts on ImageNet.
Using different lambda varies the sparsity and accuracy. Therefore, selecting appropriate $\lambda$ is important for achieving both high sparsity and high accuracy.

\subsection{Effects of \(K\) in Concept Retrieval}\label{sec:ex_concept_search}
\begin{figure}[t]
  \centering
    \centering
    \includegraphics[width=0.5\columnwidth]{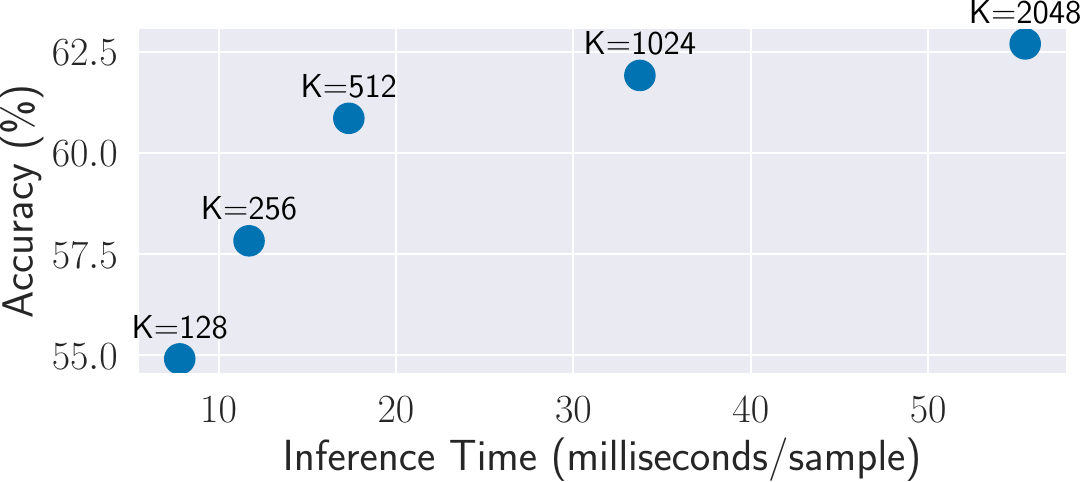}
    \captionof{figure}{Accuracy vs. inference time by varying retrieved concept number \(K\).}\label{fig:accuracy_vs_time_k}
\end{figure}

As discussed in Sec.~\ref{sec:implementation}, the retrieved concept number \(K\) in concept retrieval controls the trade-off between the accuracy and inference time.
We assess the effects of \(K\) by varying it in \([128, 256, 512, 1024, 2048]\) and measuring the top-1 accuracy and averaged inference time for processing an image.
Note that we set \(2048\) as the maximum value of \(K\) because it is the upper bound in the GPU implementation of Faiss~\citep{Johnson_IEEE19_faiss_gpu}.
Figure~\ref{fig:accuracy_vs_time_k} illustrates the relationship between the accuracy and total inference time.
As expected, the size of \(K\) produces a trade-off between accuracy and inference time.
Even so, the increase in inference time with increasing \(K\) is not explosive and is sufficiently practical since the inferences can be completed in around 55 milliseconds per sample.
The detailed breakdowns of total inference time when \(K=2048\) were 0.11 for extracting image features, 5.35 for concept retrieval, and 49.23 for concept regression, indicating that the computation time of concept regression is dominant for the total.
In future work, we explore speeding up methods for Z-CBMs to be competitive with the existing CBMs baseline that require training (e.g., Label-free CBMs, which infer a sample in 3.30 milliseconds).

\paragraph{Ethics Statement.} 
A potential ethical risk of our proposed method is the possibility that biased vocabulary contained in the concept bank may be output as explanations.
Since the concept bank is automatically generated from the caption dataset, it should be properly pre-processed using a filtering tool such as Detoxify~\citep{Detoxify} if the data source can be biased.

\paragraph{Reproducibility Statement.}
As described in Sec.~\ref{sec:implementation}~and~\ref{sec:experiment} , the implementation of the proposed method uses a publicly available code base. For example, the VLMs backbones are publicly available in the OpenAI CLIP\footnote{https://github.com/openai/CLIP} and Open CLIP\footnote{https://github.com/mlfoundations/open\_clip} GitHub repositories. All datasets are also available on the web; see the references in Sec.~\ref{sec:ex_settings} for details.
For the computation resources, we used a 24-core Intel Xeon CPU with an NVIDIA A100 GPU with 80GB VRAM.
More details of our implementation can be found in the attached code in the supplementary materials and we will make the code available on the public repository if the paper is accepted.


\end{document}